\pdfoutput=1

\documentclass[11pt,table,dvipsnames]{article}

\usepackage[preprint]{acl}

\usepackage{times}
\usepackage{latexsym}
\usepackage{graphicx}

\usepackage[T1]{fontenc}

\usepackage[utf8]{inputenc}

\usepackage{microtype}

\usepackage{inconsolata}

\usepackage{paralist}
\usepackage{inconsolata}
\usepackage{microtype}
\usepackage{CJKutf8}
\usepackage{bigstrut}
\usepackage{amsmath}
\usepackage{multirow, makecell, caption}
\usepackage{floatrow}
\newfloatcommand{capbtabbox}{table}[][\FBwidth]
\usepackage[normalem]{ulem}
\usepackage{amssymb}
\usepackage{amsfonts}
\usepackage{multicol}
\usepackage{tipa}
\usepackage{bm}
\usepackage[ruled,linesnumbered]{algorithm2e}
\usepackage{paralist}
\usepackage{IEEEtrantools}
\usepackage[switch]{lineno}
\usepackage{enumitem}
\usepackage{arydshln}
\usepackage{verbatim}
\usepackage{pifont}
\usepackage{csquotes}
\usepackage{xspace}
\usepackage{mdwlist}
\usepackage{subfigure}
\usepackage{array}
\usepackage{colortbl}
\usepackage{dsfont}
\usepackage{url}
\usepackage{booktabs}
\usepackage{tabularx}
\usepackage{bbm}
\usepackage{pgfplots}
\usepackage{tikz}
\usepackage{soul} 
\usepackage{color,xcolor} 
\usepackage{listings}
\usepackage{tcolorbox}

\author{Caiyu Hu\textsuperscript{\rm $\spadesuit$}\quad
Yikai Zhang\textsuperscript{\rm $\spadesuit$}\quad
Tinghui Zhu\textsuperscript{\rm $\spadesuit$}\quad
Yiwei Ye\textsuperscript{\rm $\diamondsuit$}\quad
Yanghua Xiao\textsuperscript{\rm $\spadesuit$}\thanks{Corresponding author.}
\\
\textsuperscript{\rm $\spadesuit$}Shanghai Key Laboratory of Data Science, School of Computer Science, Fudan University\\
\textsuperscript{\rm $\diamondsuit$}School of Computer Engineering and Science, Shanghai University\\
\texttt{\{cyhu24,ykzhang22,thzhu22\}@m.fudan.edu.cn}\\
\texttt{yiweiye@shu.edu.cn},
\texttt{shawyh@fudan.edu.cn} \\
\texttt{\url{https://caiyuhu.github.io/MCiteBench}}
}

\newcommand{\ie}{\textit{i.e.}\xspace}
\newcommand{\eg}{\textit{e.g.}\xspace}

\newcommand{\method}{\textsc{MCiteBench}\xspace}

\title{\method: A Multimodal Benchmark for Generating Text with Citations}
\setkeys{Gin}{draft=false}

\usepackage{graphicx}
\usepackage{eso-pic}

\begin{document}
\maketitle

\begin{abstract}
Multimodal Large Language Models (MLLMs) have advanced in integrating diverse modalities but frequently suffer from hallucination.
A promising solution to mitigate this issue is to generate text with citations, providing a transparent chain for verification.
However, existing work primarily focuses on generating citations for text-only content, leaving the challenges of multimodal scenarios largely unexplored.
In this paper, we introduce \method, the first benchmark designed to assess the ability of MLLMs to generate text with citations in multimodal contexts.
Our benchmark comprises data derived from academic papers and review-rebuttal interactions, featuring diverse information sources and multimodal content. 
Experimental results reveal that MLLMs struggle to ground their outputs reliably when handling multimodal input. 
Further analysis uncovers a systematic modality bias and reveals how models internally rely on different sources when generating citations, offering insights into model behavior and guiding future directions for multimodal citation tasks. 
\label{sec:abstract}
\end{abstract}

\section{Introduction}
\label{sec:intro}
Multimodal Large Language Models (MLLMs) have shown remarkable progress in integrating external information from diverse modalities, allowing them to generate responses beyond the scope of their internal knowledge~\cite{cho2024m3docrag,li2024benchmarking,zhang2024vision}.
Despite the advancements, these models frequently suffer from hallucination~\cite{huang2023survey,bai2024hallucination}, undermining the faithfulness of their outputs~\cite{zhu2024unraveling}.
A natural strategy to alleviate this issue is citation: allowing the model to attribute each generated statement to its source, thereby improving transparency and verifiability.

\begin{figure}[t]
    \centering
    \includegraphics[width=\linewidth]{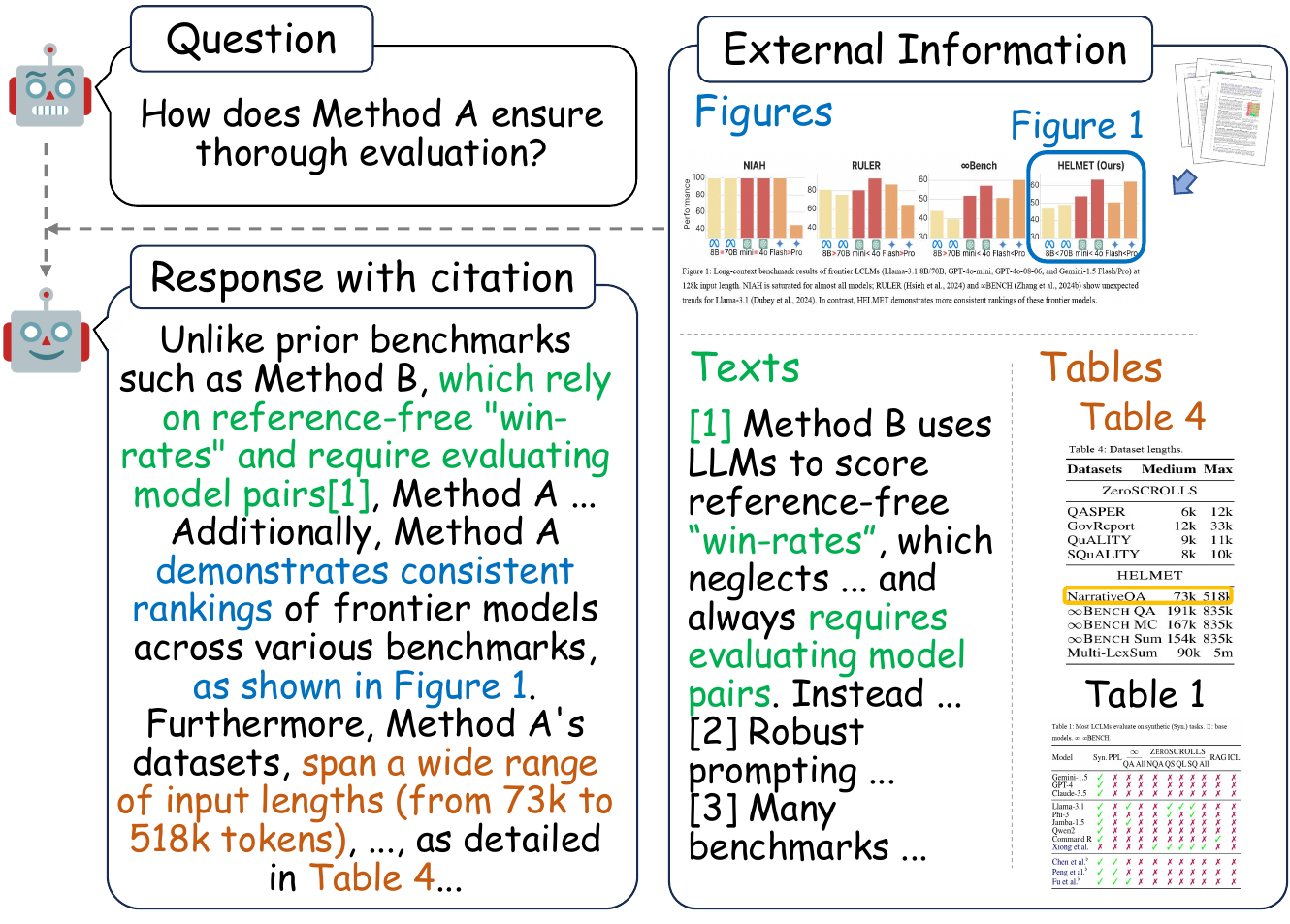}
    \caption{Illustration of the task form in \method. The model takes multimodal corpus and generates responses with explicit citations.}
    \label{fig:figure_1}
\end{figure}

Existing studies on generating text with citations mainly focus on the textual modality~\cite{gao2023enabling,liu2023evaluating}.
However, real-world information sources are inherently multimodal, often conveying information that cannot be captured by text alone.
Although common in practice, citations from non-textual modalities remain underexplored.
Grounding model responses in multimodal sources can improve faithfulness and quality (see Figure~\ref{fig:figure_1}).
At the same time, this task poses several challenges for MLLMs.
The model must understand cross-modal content, assess the sufficiency of evidence, and remain robust to irrelevant or distracting input.
These challenges are still under exploration.
In this paper, we construct a benchmark to systematically evaluate MLLMs in generating text with citation from multimodal input.

However, building such a benchmark is challenging.
First, constructing high-quality question-answer data with multimodal evidence is non-trivial.
It requires not only the accurate extraction of heterogeneous evidence (e.g., tables, figures, and passages), but also careful alignment between the evidence and the answer.
In cases where multiple pieces of evidence jointly support an answer, it is critical to ensure their mutual consistency and sufficiency.
Second, evaluating MLLMs in this setting introduces additional complexity.
A key issue is how to assess cross-modal entailment—whether the cited evidence truly supports the generated answer.
Moreover, the citation must correspond closely to the response, ensuring that the retrieved evidence is both necessary and relevant to the output.
These challenges highlight the need for a comprehensive evaluation framework that examines multiple dimensions of model performance.

In this paper, we propose \method, the first benchmark for evaluating the ability of MLLMs to generate text with citations in multimodal settings.
To address the challenges outlined above, we begin by collecting academic papers and extracting reliable information sources across multiple modalities.
These sources are rigorously filtered to form a high-quality attribution corpus.
Based on this corpus, we construct question–answer pairs using review–rebuttal interactions, where each answer is supported by evidence.
To comprehensively evaluate model performance, we assess models along three axes: citation quality, source reliability, and answer accuracy.
Extensive experiments reveal several notable findings:
\begin{inparaenum}[1)]
    \item While MLLMs can often answer questions correctly, they struggle to generate accurate citations, particularly when the evidence spans multiple sources.
    \item MLLMs are better at attributing citations to textual than to visual evidence, suggesting a potential modality bias.
\end{inparaenum}

Our contributions are summarized as follows:
\begin{itemize}
\item To the best of our knowledge, \method{} is the first benchmark that systematically evaluates the ability of MLLMs to generate text with citations from multimodal input.
\item \method comprises 3,000 samples of different difficulty levels, including both single- and multi-source evidence, as well as single- and mixed-modality cases.
To support comprehensive evaluation, we define multi-dimensional metrics capturing citation quality, source reliability, and answer accuracy.
\item We conduct experiments to assess the models’ ability to generate text with citations across different modalities. Results reveal that MLLMs exhibit a modality bias, favoring textual over visual sources in citation generation.
\end{itemize}

\section{Related Work}
\label{sec:related}
\paragraph{Generating Text with Citations}
Recent efforts have explored the task of generating text with citations, where models are required to produce responses with explicit references to supporting sources.
\citet{gao2023enabling,liu2023evaluating} first introduced this setting to improve the verifiability of model responses.
Subsequent works have explored two main paradigms: generating both the response and citations simultaneously~\cite{aly2024learning,huang2024training}, and attaching citations in a post-processing step~\cite{slobodkin2024attribute,li2024citation}.
These approaches have also been extended to tasks such as long-context citation~\cite{zhang2024longcite} and fine-grained attribution~\cite{xu2024aliice}.
Another related line of work is traditional citation text generation, which typically refers to generating citation sentences in academic papers that contain specific scientific claims and cite prior work~\cite{li2024related,mandal2024contextualizing,csahinucc2024systematic}.
However, existing studies focus almost exclusively on textual evidence, limiting their applicability in real-world multimodal scenarios.
In this work, we address this gap by incorporating figure and tabular content as citation sources and evaluating model attribution in multimodal contexts.

\begin{figure*}[!h]
    \centering
    \includegraphics[width=\linewidth]{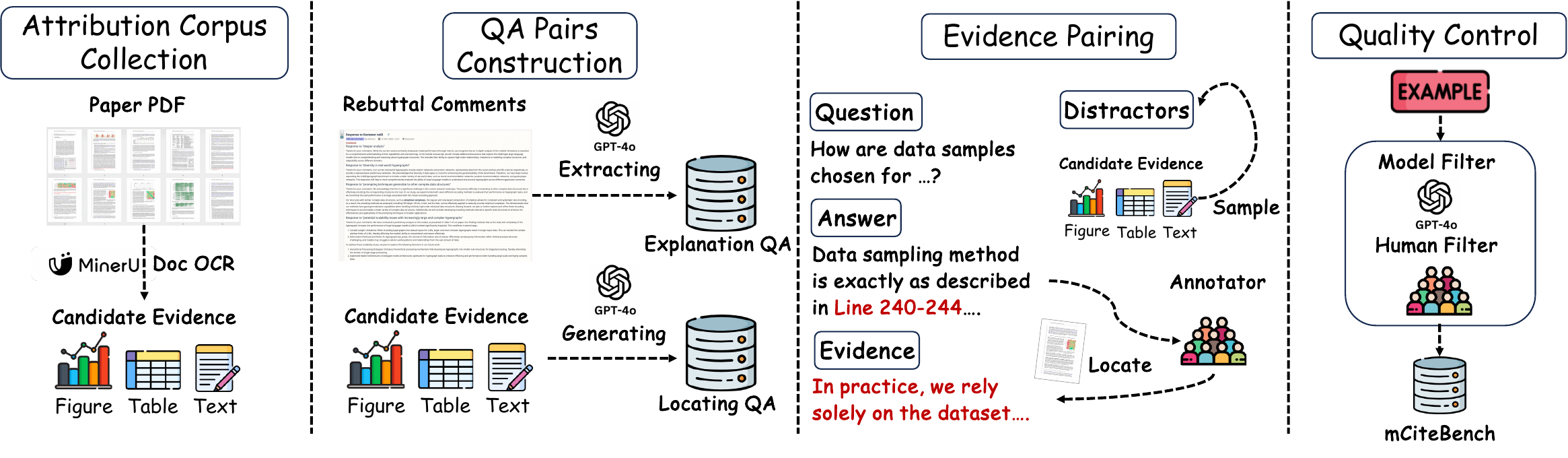}
    \caption{The construction pipeline of \method. Initially, we collect multimodal academic papers along with their corresponding review-rebuttal interactions and then parse the papers to extract candidate evidence. GPT-4o is used to extract explanation QA pairs from the comments and generate locating QA pairs. Next, human annotators match the references in the answers to the relevant content in the original papers. Finally, the data filtered and labeled by the model is manually verified by human annotators to ensure consistency and accuracy.}
    \label{fig:figure_2}
\end{figure*}

\paragraph{Multimodal RAG}
Multimodal retrieval-augmented generation (mRAG)~\cite{zhao2023retrieving} augments multimodal large language models with retrieved external information, enabling them to answer queries that cannot be resolved using internal knowledge alone.
\citet{zhang2024vision} acquire unknown visual knowledge through web search to aid in answering queries, while~\citet{li2024benchmarking} builds a self-adaptive retrieval agent to plan the reasoning path.
Additionally,~\citet{cho2024m3docrag} improve multi-page and multi-document understanding through multimodal retrieval.
While these approaches integrate retrieval into the generation pipeline, they do not assess whether the generated responses faithfully reflect the retrieved content.
In this work, we shift the focus from retrieval itself to attribution: evaluating whether the model can correctly ground its outputs in the provided multimodal sources.

\section{\method}
\label{sec:method}
In this section, we define the task of generating text with citations from multimodal input and describe the construction of our benchmark, \method. As shown in Figure~\ref{fig:figure_2}, the pipeline consists of four main stages: \textbf{Attribution Corpus Collection}, \textbf{QA Pairs Construction}, \textbf{Evidence Pairing}, and \textbf{Quality Control}.
We begin by collecting academic papers, which serve as a source of rich multimodal content. 
Based on these papers, we construct question–answer pairs from review–rebuttal interactions. 
Human annotators are employed to link answers to their supporting evidence. 

\subsection{Task Definition}
Given a query \(q\) and a multimodal evidence set \(M\), where \(M\) includes both the ground truth evidence and distractors related to \(q\), the model is required to generate an answer \(a\) along with a set of citations \(C\).
For each sentence \(s_i\) in the answer, the model generates a set of citations \(C_i = \{ c_{i,1}, c_{i,2}, \dots, c_{i,k_i} \}\), where \(k_i\) denotes the number of cited evidence associated with sentence \(s_i\). 
Each citation \(c_{i,j}\) refers to a specific piece of evidence from the multimodal evidence set \(M\).

\subsection{Attribution Corpus Collection}
To evaluate how well MLLMs generate text with citations, an attribution corpus that includes multimodal information sources and allows for easy verification of cited evidence is needed.
In \method, we use academic papers as the attribution corpus because of the following characteristics:
\begin{inparaenum}[1)]
\item Academic papers contain rich content from multiple modalities (\eg, text, figure, and table) that individually or collectively support the arguments.
\item The information sources in academic papers are numbered (\eg, ``Figure 1'', ``Table 2'', and text in ``Line 10''), making it easy to match them with the cited results. 
\item Academic papers cover the latest contents beyond pre-training data, reducing the risk of data leakage.
\end{inparaenum}

We collect papers from OpenReview and extract multimodal content using MinerU~\cite{wang2024mineru}, a state-of-the-art document parsing framework.
To avoid contamination from model training data, we focus on ICLR 2025 submissions, which became publicly available in November 2024—after the knowledge cutoff of the evaluated models.
ICLR is chosen for its open review process, which includes accessible reviews and author responses, offering reliable structure for citation annotation.
From this collection, we obtain a diverse set of multimodal content, including over 400k text paragraphs, 40k images, and 9k tables, which serve as candidate evidence.
A subset of this corpus is selected as candidate evidence and distractors for constructing the final 3k evaluation samples.

\subsection{QA Pairs Construction}
After collecting the attribution corpus, we construct question–answer pairs with explicit references to the supporting evidence.
Establishing a reliable correlation between questions and evidence is challenging, as the source of information must be accurately linked to the generated answers.

We divide \method data into two categories: \textbf{Explanation} and \textbf{Locating}. 
Explanation questions require in-depth analysis of evidence and often yield long-form responses (\eg, ``How is the model’s performance evaluated?'').
In contrast, Locating questions are straightforward and can be answered by directly identifying the correct evidence (\eg, ``Which model performs better on the XYZ benchmark, GPT-4o or GPT-4o-mini?'').

For Locating questions, we use GPT-4o to generate structured QA pairs with supporting details. Specifically, we construct QA pairs \(( Q, A)\), where each question \(q_i \in Q\) is formulated based on specific evidence, and each answer \(a_i \in A\) is directly linked to the corresponding source.

However, generating questions that require information from multiple sources remains a challenge for MLLMs.
Models often fail to integrate all necessary evidence, resulting in questions that can be answered by a single source rather than all selected evidence.
To address this, we leverage review-rebuttal interactions to construct Explanation QA pairs.
In this setting, reviewers’ questions and authors’ responses are used, with responses grounded in multiple evidence segments from the paper (\ie, attribution corpus).
From these data, we construct QA pairs \(( Q, A) \) by extracting questions \(q_i\) and the corresponding answers \(a_i\).\footnote{Details of prompt design and reference extraction strategies are in Appendix~\ref{app:data_processing_prompt}}

\subsection{Evidence Pairing}
Review–rebuttal interactions often include rich evidence in the authors’ responses to support their claims. For example, when addressing a reviewer’s concern about model performance, an author might respond, \textit{“Our approach achieves 85.2\% accuracy, as shown in Table 3 and discussed in Section 4.2.”}
These references provide valuable entry points for identifying the evidence that grounds the answer.
Therefore, we extract the supportive evidence \(e_i \in E\) from \(a_i \in A\) to construct \( ( Q, A, E ) \) triplets.
While \( E \) provides explicit references (e.g., ``Table 3'', ``Section 4.2''), these references must be resolved to their corresponding content in the source papers before they can be used as input for MLLMs.
To achieve this, human annotators manually map each reference to the associated content in the original paper, categorizing the evidence as either text, image, or table.

\paragraph{Distractor Construction.}
To evaluate whether models can correctly cite relevant sources while ignoring irrelevant ones, we introduce distractor content into the input.
These distractors are sampled from the same paper, ensuring a balanced distribution of multimodal content (text, images, tables). 
Each final sample in \method is formatted as \((Q, A, E, D)\), where \(Q\) is the question, \(A\) is the correct answer, \(E\) is the evidence and \(D\) is the distractors.

\begin{table}[t]
  \centering
  \small
  \begin{tabular}{lc}
    \toprule
    \textbf{Statistic}
                               & \textbf{Number} \\
    \midrule
    \textbf{Total questions}   & 3,000           \\
    - Explanation                & 2,000           \\
    - Locating               & 1,000           \\
    \midrule
    \textbf{Evidence sources}    &                 \\
    - Single-source              & 2,538            \\
    - Multi-source               & 462             \\
    \midrule
    \textbf{Evidence modality}    &                 \\
    - Text                       & 1,243            \\
    - Figure                     & 941             \\
    - Table                      & 533             \\
    - Mixed                      & 283             \\
    \midrule
    Total papers                & 1,749            \\
    Average questions per paper & 1.72            \\
    \bottomrule
  \end{tabular}
  \caption{Statistics of \method.}
  \label{tab:data_statistics}
\end{table}


\subsection{Quality Control}
After constructing \((Q, A, E, D)\), we apply a quality control pipeline that first uses automated filtering followed by human verification.
Initially, GPT-4o assigns quality labels and filters out low-quality samples based on predefined criteria such as relevance, clarity, and evidence alignment. 
The filtered candidates are then manually verified by annotators to ensure consistency and accuracy, focusing on removing any unclear or incorrect instances.\footnote{Details of the human annotation process can be found in Appendix~\ref{app:quality_control}.}

\subsection{Statistics of \method}
As shown in Table~\ref{tab:data_statistics}, \method comprises 3,000 data samples for evaluating the ability of MLLMs to generate text with citations, extracted from 1,749 academic papers with an average of 1.72 questions per paper.
Among these, 2,000 are Explanation tasks that require detailed evidence analysis and often lead to long-form answers, while 1,000 are Locating tasks that focus on direct evidence identification.
The evidence is balanced across modalities, with 1,243 textual, 1,474 visual (including 941 figures and 533 tables), and 283 mixed-modality sources, ensuring diverse multimodal attribution scenarios.

\section{Evaluation Metrics}
\label{sec:evaluation}
We evaluate the models across three dimensions: \textbf{citation quality}, \textbf{source reliability}, and \textbf{answer accuracy}. Using \textbf{Citation F1}, we assess whether the cited evidence accurately and sufficiently supports the model’s response. 
Source reliability ensures that the model’s response cites the ground truth source needed to answer the query. 
We measure this by comparing the model-generated citation with ground truth citation, using both \textbf{Source F1} and \textbf{Source Exact Match} scores. 
Answer accuracy metrics are designed to assess whether the model’s response correctly addresses the query.

\begin{figure}[t]
    \centering
    \includegraphics[width=\linewidth]{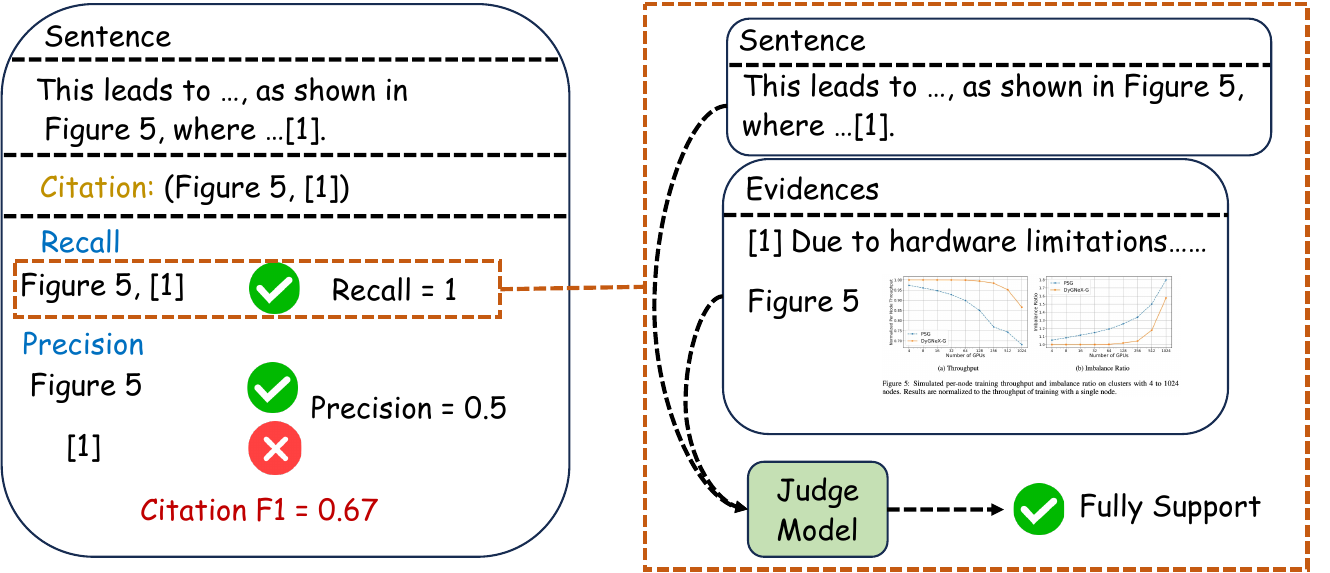}
    \caption{The calculation of Citation F1.}
    \label{fig:citation_quality_eval}
\end{figure}

\paragraph{Citation F1 (C-F1).}
Citation quality is evaluated using Citation F1, which measures the alignment between cited evidence and the generated response, ensuring that the response is supported by the cited evidence without including irrelevant ones. 

As illustrated in Figure \ref{fig:citation_quality_eval}, a judge model evaluates whether each sentence is supported by its cited evidence.
Citation Recall is calculated using a scoring system inspired by LongCite~\cite{zhang2024longcite}, categorizing citations into three levels: No support, Partially supported, and Fully supported, with corresponding scores of 0, 0.5, and 1.
Citation Precision is determined on a binary scale, scored as either relevant (1) or irrelevant (0) to the cited evidence.
For sentences citing multiple sources, the final precision score is the average across all cited evidence.
Finally, Citation F1 is computed as the harmonic mean of Recall and Precision, providing a balanced measure of the model’s citation quality.

\newcolumntype{B}{>{\columncolor{blue!4}}c}
\newcolumntype{d}{>{\columncolor{brown!4}}c}
\newcolumntype{q}{>{\columncolor{green!4}}c}
\newcolumntype{E}{>{\columncolor{purple!4}}c}

\makeatletter
\def\adl@drawiv#1#2#3{%
        \hskip.5\tabcolsep
        \xleaders#3{#2.5\@tempdimb #1{1}#2.5\@tempdimb}%
                #2\z@ plus1fil minus1fil\relax
        \hskip.5\tabcolsep}
\newcommand{\cdashlinelr}[1]{%
  \noalign{\vskip\aboverulesep
           \global\let\@dashdrawstore\adl@draw
           \global\let\adl@draw\adl@drawiv}
  \cdashline{#1}
  \noalign{\global\let\adl@draw\@dashdrawstore
           \vskip\belowrulesep}}
\makeatother

\setlength\tabcolsep{4pt}
\begin{table*}[!h]
  \centering
  \small
  \begin{tabular}{lBBBBqqqqEEEE}
    \toprule
    \multirow{3}{*}[-1.5ex]{\textbf{Models}}
      & \multicolumn{8}{c}{\textbf{Explanation}}
      & \multicolumn{4}{c}{\textbf{Locating}}     \\
    \cmidrule(lr){2-9} \cmidrule(lr){10-13}
      & \multicolumn{4}{c}{\textbf{Single-Source}}
      & \multicolumn{4}{c}{\textbf{Multi-Source}}
      & \multicolumn{4}{c}{\textbf{Single-Source}} \\
    \cmidrule(lr){2-5}  \cmidrule(lr){6-9} \cmidrule(lr){10-13}
      & \multicolumn{1}{c}{\textbf{C-F1}}
      & \multicolumn{1}{c}{\textbf{S-F1}}
      & \multicolumn{1}{c}{\textbf{S-EM}}
      & \multicolumn{1}{c}{\textbf{Acc}}
      & \multicolumn{1}{c}{\textbf{C-F1}}
      & \multicolumn{1}{c}{\textbf{S-F1}}
      & \multicolumn{1}{c}{\textbf{S-EM}}
      & \multicolumn{1}{c}{\textbf{Acc}}
      & \multicolumn{1}{c}{\textbf{C-F1}}
      & \multicolumn{1}{c}{\textbf{S-F1}}
      & \multicolumn{1}{c}{\textbf{S-EM}}
      & \multicolumn{1}{c}{\textbf{Acc}}          \\
    \midrule
    \multicolumn{13}{>{\columncolor{gray!10}}c}{{\textit{Open-Source Models (7-14B)}}} \\
    LLaVA-OV-7B         & 19.93 & 10.84 & 5.34 & 47.79 & 31.14 & 22.48 & 1.26 & 49.68 & 26.31 & 20.93 & 11.63 & 60.10 \\
    LLaVA-OV-7B-Chat    & 28.77 & 13.90 & 1.43 & 47.76 & 35.74 & 29.82 & 3.00 & 49.78 & 29.58 & 23.33 & 4.05 & 53.85  \\
    MiniCPM-V-2.6   & 49.12 & 35.23 & 22.81 & 51.30 & 57.90 & 41.74 & 5.88 & 52.60 & 47.93 & 52.73 & 42.94 & 83.55  \\
    Qwen2-VL-7B      & 58.46 & 42.98 & \textbf{35.36} & 51.59 & 58.64 & 36.62 & 2.36 & 53.03 & 53.99 & 54.71 & 46.32 & 87.45 \\
    InternVL2.5-8B  & \underline{58.47} & \underline{45.13} & \underline{33.45} & 51.53 & 63.97 & 45.50 & 9.86 & 52.92 & 55.94 & 64.17 & 56.33 & 83.90 \\
    Llama-3.2-Vision-11B        & 19.65 & 14.06 & 9.60 & 48.63 & 31.16 & 25.87 & 1.22 & 49.35 & 26.56 & 16.56 & 11.80 & 61.40   \\
    \cdashlinelr{2-13}
    \multicolumn{13}{>{\columncolor{gray!10}}c}{\textit{Open-Source Models (>70B)}} \\
    Qwen2-VL-72B     & 53.60 & 44.81 & 32.01 & \underline{52.60} & 64.66 & 50.53 & 8.96 & 52.38 & \underline{58.75} & \underline{68.86} & \underline{61.48} & \underline{90.25} \\
    InternVL2.5-78B & 54.52 & 42.44 & 25.40 & 52.34 & \underline{71.03} & \underline{57.65} & \underline{16.86} & \underline{54.87} & 50.57 & 57.60 & 52.20 & 90.10 \\
    Llama-3.2-Vision-90B    & 35.33 & 28.05 & 12.30 & 50.00 & 46.08 & 46.73 & 10.35 & 51.41 & 43.69 & 49.07 & 32.83 & 74.75 \\
    \cdashlinelr{2-13}
    \multicolumn{13}{>{\columncolor{gray!10}}c}{\textit{Proprietary Models}} \\
    GPT-4o-mini      & 43.99 & 34.42 & 15.48 & 52.08 & 57.81 & 50.22 & 8.39 & 54.22 & 53.71 & 58.57 & 46.56 & 88.50 \\
    GPT-4o           & \textbf{84.24} & \textbf{56.82} & 24.50 & \textbf{54.32} & \textbf{89.19} & \textbf{67.56} & \textbf{21.27} & \textbf{56.60} & \textbf{91.45} & \textbf{85.74} & \textbf{69.45} & \textbf{90.45} \\
    \bottomrule
  \end{tabular}
  \caption{Main results on \method. The highest score is highlighted in \textbf{bold}, and the second highest score is \underline{underlined}. C-F1, S-F1, and S-EM represent Citation F1, Source F1, and Source Exact Match scores, respectively. Acc stands for Accuracy.}
  \label{tab:main_exp}
\end{table*}

\begin{figure}[t]
    \centering
    \includegraphics[width=\linewidth]{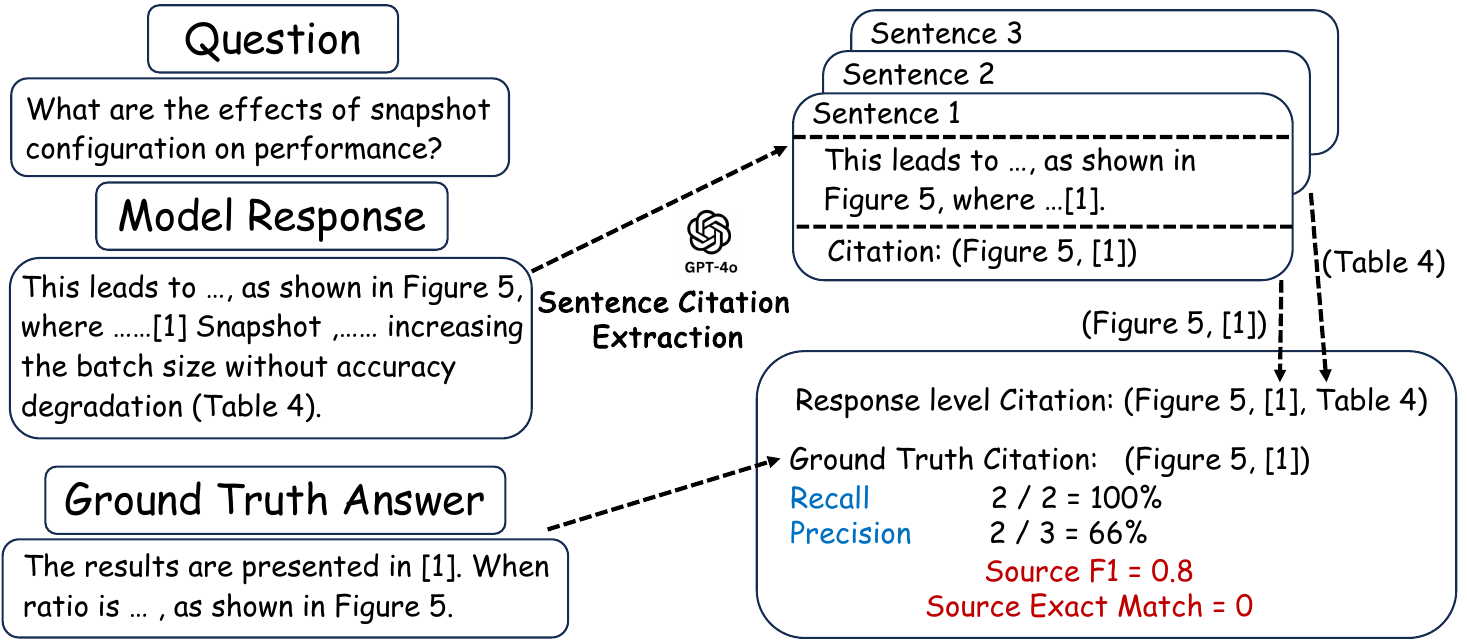}
    \caption{The calculation of Source F1 and Source Exact Match.}
    \label{fig:source_reliability_eval}
\end{figure}

\paragraph{Source F1 (S-F1).}
As shown in Figure \ref{fig:source_reliability_eval}, Source F1 measures the alignment between citations in the model’s response and ground truth citations, evaluating whether the model cites evidence that aids in answering the query.

We first split the model-generated responses into sentence-citation pairs \((s_i, c_i)\) using GPT-4o.
These sentence-level citations are then aggregated to form response-level citations, which are compared against the ground truth. The precision, recall, and F1 score are calculated as follows:

{
\small
\begin{equation}
\text{Source Precision} = \frac{|C_{\text{pred}} \cap C_{\text{gt}}|}{|C_{\text{pred}}|},
\end{equation}
}

{
\small
\begin{equation}
\text{Source Recall} = \frac{|C_{\text{pred}} \cap C_{\text{gt}}|}{|C_{\text{gt}}|},
\end{equation}
}

We calculate Source F1 by computing the harmonic mean of Recall and Precision.
\(C_{\text{pred}}\) represents the set of citations generated by the model, and \(C_{\text{gt}}\) denotes the ground truth citations. The intersection \(C_{\text{pred}} \cap C_{\text{gt}}\) counts the correctly cited evidence.

\paragraph{Source Exact Match (S-EM).}
The Source Exact Match metric provides a stricter evaluation, indicating whether the model’s response-level citation is the same as the ground truth.

{
\small
\begin{equation}
\text{Source EM} =
\begin{cases} 
1, & \text{if } C_{\text{pred}} = C_{\text{gt}} \\
0, & \text{otherwise}
\end{cases}
\end{equation}
}

\paragraph{Accuracy (Acc).}
We evaluate answer accuracy using the LLM-As-Judge~\cite{zheng2023judging,liu2023alignbench} framework for both Explanation and Locating questions.
The judge model scores each response and reference answer according to criteria specific to each question type, and the scores are then normalized.
In Explanation cases, direct comparison with a ground truth answer is not feasible. 
Instead, we use the authors’ responses as the reference and employ a judge model to evaluate the generated answers based on their relevance, logical consistency, and fluency.
In Locating scenarios, this evaluation method mitigates issues related to errors caused by minor formatting differences.\footnote{Detailed scoring criteria and judgment prompts are provided in the Appendix~\ref{app:evaluation_metric_prompt}.}

\section{Experiments}
\label{sec:experiment}
\subsection{Evaluation Settings}

\paragraph{Implement Details.}
In this work, we indicate citations from textual content using box brackets (\eg, ``[1]''), and refer to figures and tables by the indices in their captions (\eg, ``Figure 3'', ``Table 2''). We conduct an ablation study to assess the impact of including figure captions in the input.\footnote{See Table~\ref{tab:caption_ablation} for details in Appendix~\ref{app:caption_ablation}.}
For both single-source and multi-source evidence questions, the multimodal corpus \(M\) comprises 5 items, including the ground truth evidence and distractors.
Distractors are randomly selected from other content within the same paper.

\paragraph{Judge Model.}
In this study, we use GPT-4o to assess the entailment relationship between model responses and their cited evidence.\footnote{We validate GPT-4o’s reliability in Appendix~\ref{app:human_check}, and further verify in Appendix~\ref{app:llm_judge} that it does not exhibit strong self-preference when evaluating responses in our task.}

\paragraph{Model Choice.}
For open-source models, we test InternVL-2.5 (8B/78B)~\cite{chen2024expanding}, Qwen2-VL (7B/78B)~\cite{wang2024qwen2}, Llama 3.2-Vision (11B/90B)~\cite{meta2024llama}, Llava-OneVision (and its chat version) and MiniCPM-V-2.6~\cite{yao2024minicpm}.
For proprietary models, we test GPT-4o (GPT-4o-2024-11-20) and GPT-4o-mini (GPT-4o-mini-2024-07-18)~\cite{hurst2024gpt}.

\subsection{Main Results}
As shown in Table~\ref{tab:main_exp}, smaller open-source models achieve lower Citation F1 scores and struggle to select evidence that adequately supports their responses.
Furthermore, they also perform poorly in selecting evidence that directly answers the query, as shown by their low Source F1 and Source Exact Match scores.
As model size increases, we observe an improvement in citation performance, suggesting that scaling model size enhances attribution capability.
In comparison, GPT-4o achieves an 84.24\% Citation F1 score on single-source Explanation questions, demonstrating strong citation quality. 
However, it struggles with source reliability, with Source Exact Match scores remaining low at 24.50\% for single-source and 21.27\% for multi-source settings.
This indicates that even state-of-the-art models struggle to consistently cite evidence that is directly relevant to answering the query, underscoring the difficulty of precise citation in multimodal contexts.

\paragraph{Does Question Difficulty Influence Model Citation Performance?}
Model performance reflects the difficulty of the questions, with higher accuracy scores observed on locating questions compared to explanation questions, indicating that explanation tasks are more challenging. 
As shown in Table~\ref{tab:main_exp}, as question difficulty increases, model citation performance tends to decrease.
For instance, GPT-4o achieves 85.74\% in Source F1 for single source locating questions but drops to 56.82\% for single source explanation questions. 
Explanation questions place higher demands on citation generation, as they require an in-depth analysis of the inputs.

\paragraph{How Do Multi-Source Scenarios Affect Generating Text with Citations in MLLMs?}
In multi-source settings, models tend to achieve higher Citation F1 and Source F1 scores, as multiple valid references allow for partial credit. 
Unlike single-source questions with only one correct citation, multi-source questions permit credit for correctly identifying any subset of the ground truth, naturally resulting in higher metric values.
However, the stricter Source Exact Match metric is lower than in single-source scenarios.
This highlights the challenge of citing in multi-source scenarios, where models must correctly include relevant sources while avoiding irrelevant ones.

\subsection{Analysis}
In this section, we discuss several research questions, revealing the inherent biases in the task.

\paragraph{RQ1: Can MLLMs Accurately Identify the Source Needed to Answer a Question?}

\begin{table}[t]
  \centering
  \small
  \begin{tabular}{lcccc}
    \toprule
    \multirow{2}{*}[-1ex]{\textbf{Model}} & \multirow{2}{*}[-1ex]{\textbf{Overall}} & \multicolumn{3}{c}{\textbf{By Modality}} \\
    \cmidrule(lr){3-5}
                  &                  & \textbf{Figure} & \textbf{Table} & \textbf{Text} \\
    \midrule
    \textcolor{gray!50}{\textbf{Open-Source(7-14B)}} \\
    Qwen2-VL-7B-Instruct & 0.45 & 0.40 & 0.38 & 0.55 \\
    InternVL2\_5-8B & 0.48 & 0.37 & 0.42 & 0.65 \\
    \textcolor{gray!50}{\textbf{Open-Source(>70B)}} \\
    Qwen2-VL-72B-Instruct & 0.59 & 0.50 & 0.57 & 0.71 \\
    InternVL2\_5-78B & 0.58 & 0.51 & 0.50 & 0.72 \\
    
    \textcolor{gray!50}{\textbf{Proprietary}} \\
    gpt-4o-mini & 0.52 & 0.47 & 0.48 & 0.61 \\
    gpt-4o-2024-11-20 & 0.60 & 0.52 & 0.55 & 0.73 \\
    \bottomrule
  \end{tabular}
  \caption{Model accuracy on identifying the most relevant source for answering a question under the multi-choice setting.}
  \label{tab:qa_modality_results}
\end{table}

Generating text with citation can be abstracted into a two-stage process:
(1) generating a response, and
(2) mapping that response to the appropriate supporting input sources by producing attribution tokens such as “[1]” or “Figure 3”.

Instead of requiring the model to generate an answer and then attribute it, we directly evaluate its ability to identify which source would be most helpful in answering a given question. Specifically, we ask:~\textit{Can a model identify the correct source needed to answer a given question?}

\vspace{-3pt}
\paragraph{Settings}
We construct a probing task based on Single-Source Explanation QA. For each example, we provide the model with a question and 5 candidate sources (1 correct + 4 distractors). The model is tasked with selecting which source would be most helpful in answering the question.\footnote{See Table~\ref{prompt:source_attribution_mcq} for details in Appendix~\ref{app:source_identification_prompt}.}

\vspace{-3pt}
\paragraph{Results}
Results are presented in Table~\ref{tab:qa_modality_results}. Importantly, this task directly evaluates the model’s ability to identify relevant sources based solely on the question, rather than relying on model-generated answers or intermediate claims. 
Despite this seemingly simplified setting, no model achieves more than 60\% accuracy, highlighting the persistent difficulty in accurately grounding questions in the correct source.

In addition, we observe a consistent performance gap across modalities: models perform better when reasoning over textual sources compared to visual inputs such as figures and tables, which leads to our next research question.

\paragraph{RQ2: Does Modality Influence Citation Performance?}

\begin{figure}[t]
    \centering
    \includegraphics[width=\linewidth]{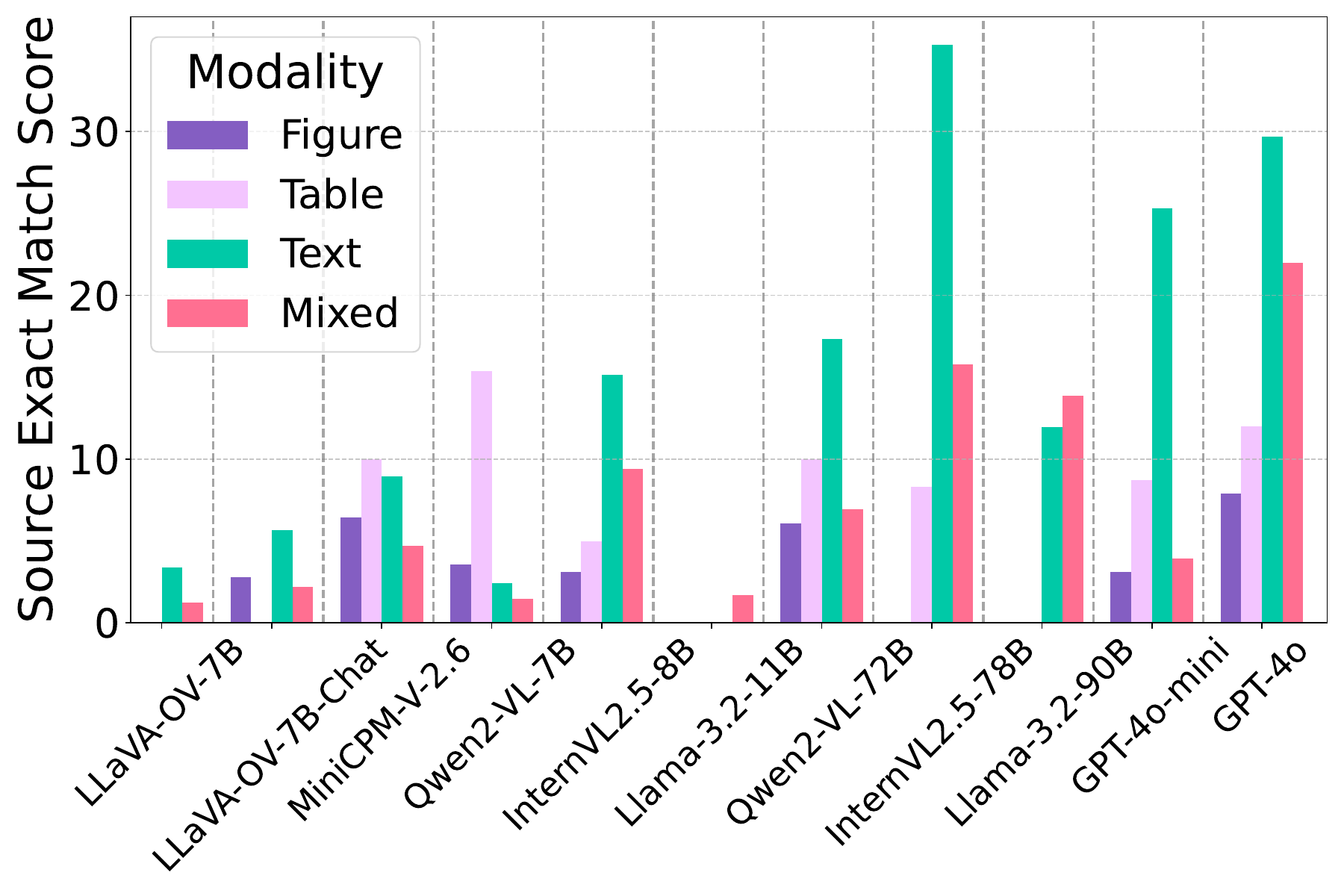}
    \caption{Source Exact Match score of models on the \method benchmark across different modalities, under the multi-source explanation setting with two gold evidence items per question.}
    \label{fig:modality_label_em_score}
\end{figure}

\begin{figure}[!t]
    \centering    
    \includegraphics[width=\linewidth]{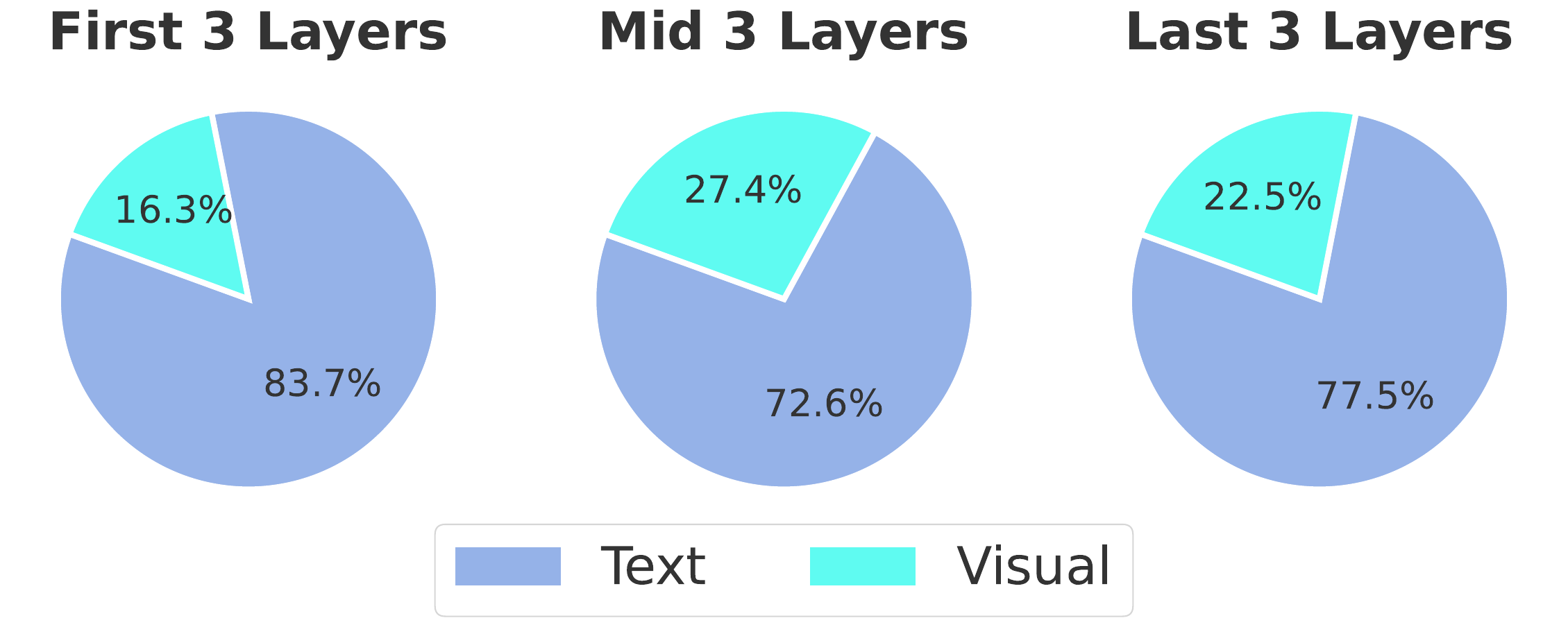}
    \caption{Attention distribution across multimodal sources.}
    \label{fig:modality_attention}
\end{figure}

We analyze model performance in instances where the evidence modality comes from mixed modalities. The number of evidence is set to 2, and we compare this with data from single modalities with the same number of evidence pieces.
As shown in Figure~\ref{fig:modality_label_em_score}, most models achieve high Source EM scores when the ground truth evidence is textual but perform poorly when it is visual.
This suggests that although MLLMs can process multimodal inputs, they are better at aligning with textual evidence than accurately citing visual information when generating responses.

\begin{figure*}[!t]
    \centering
    \includegraphics[width=0.90\linewidth]{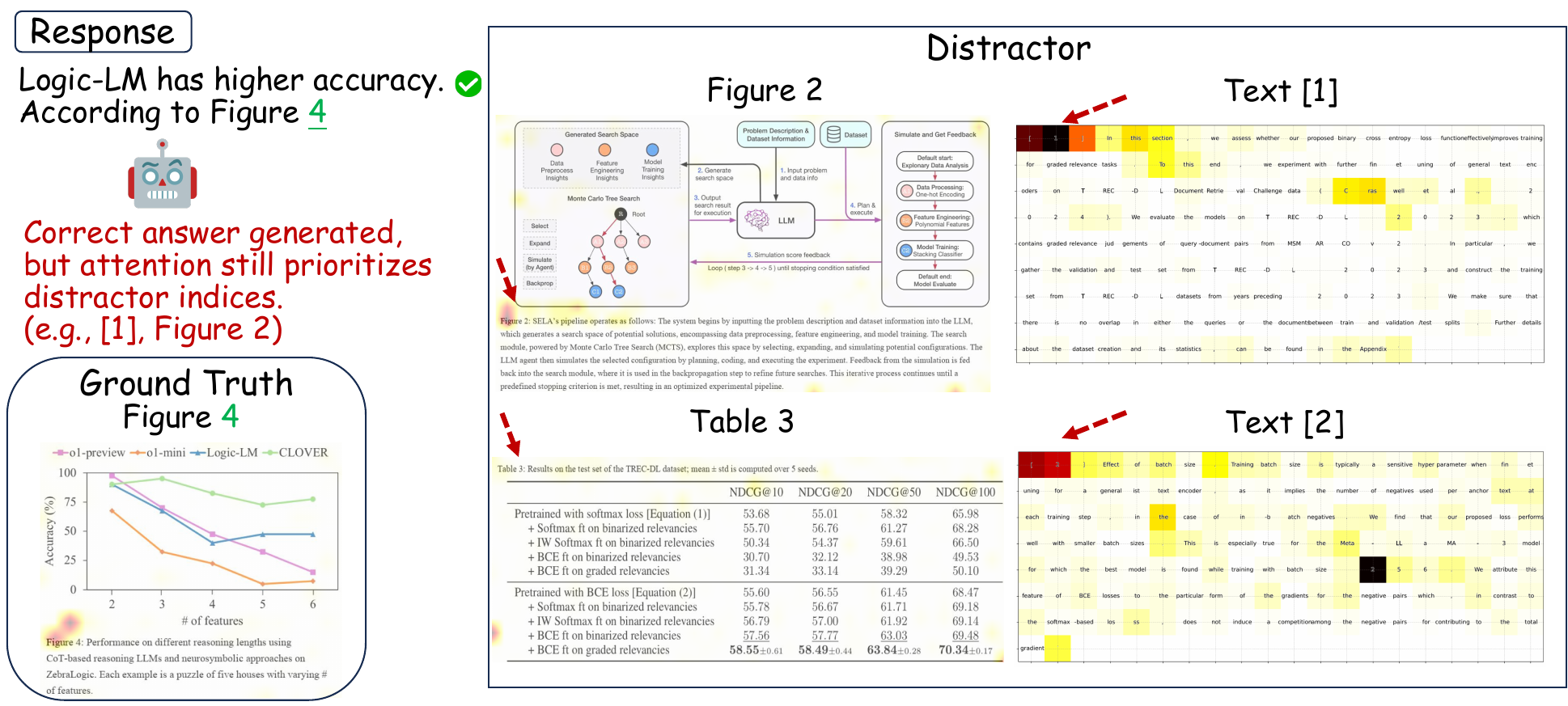}
    \caption{Attention heatmap during source reference generation. The heatmap shows how the model distributes attention when generating the next token in its response, continuing the sentence \textit{``Logic-LM has higher accuracy. According to Figure \uline{$\Diamond$}’’}. Although the model answers correctly, its attention in the distractors remains focused on index positions (\eg, ``[1]'', ``Figure 2'').}
    \label{fig:attention_heatmap}
\end{figure*}

To further investigate this, we analyze MLLMs' attention patterns when processing mixed-modality inputs. 
Using Qwen2-VL-7B as the test model, we calculate the attention distribution across multimodal inputs by averaging attention head scores and normalizing by input source token length across different layers.
As shown in Figure \ref{fig:modality_attention}, the model allocates fewer attention scores to visual inputs compared to text.
In contrast, textual information maintains consistently high attention throughout, with 83.7\% in early layers and 77.5\% in later layers.
This indicates that while the model processes all modalities, it prioritizes textual content and utilizes it more effectively than visual data.

\paragraph{RQ3: What Do Models Look At When Generating Citations?}
\label{RQ:Positional_Tokens}

Correctly generating source-identifying tokens (e.g., ``[1]'', ``Figure 2'') leads to better performance and higher attribution scores.
To better understand how models process ground truth evidence and distractors, we analyze their attention distribution when generating source-identifying tokens.

\vspace{-3pt}
\paragraph{Settings}
Specifically, we examine the attention patterns of Qwen2-VL-7B when continuing a partially generated sentence ending in \textit{``According to Figure''}, and tasked with predicting the next token (e.g., ``4''). This allows us to assess which input regions the model attends to when making source attribution decisions.

Specifically, we focus on its behavior when predicting the next token after \textit{``According to Figure \uline{$\Diamond$}''} in its response.
Notably, the distractors are sampled from unrelated papers, meaning they provide no useful information for answering the question.

\vspace{-3pt}
\paragraph{Results}
As shown in Figure~\ref{fig:attention_heatmap}, the model’s attention heatmap reveals an intriguing pattern: even when the response is based entirely on a specific piece of evidence, the model’s attention does not solely focus on it.
When generating the token after ``According to Figure'', the model’s attention remains high on textual index positions (\eg, ``[1]'', ``[2]''), even though the context suggests the model should focus on figure evidence.
This suggests that while the model correctly cites the source, it maintains a broader contextual awareness by attending to multiple potential evidence.

\section{Conclusion}
\label{sec:conclusion}
In this paper, we introduce \method, a high-quality benchmark built from academic papers and their review–rebuttal interactions, to evaluate the ability of MLLMs to generate text with citations from multimodal input.
Leveraging this benchmark, we conduct a detailed evaluation of model performance across multiple dimensions.
Through extensive experiments, we find that existing models struggle to accurately attribute their outputs to the correct multimodal sources. 
Furthermore, we dive deep into the analysis of attention distribution during citation generation and uncover modality bias exhibited by current models.
We hope that \method offers valuable insights into generating text with citations and contributes to the development of models capable of producing faithful and verifiable responses.

\section*{Limitations}
In \method, we construct multi-level questions and build an evaluation pipeline for multimodal inputs. 
However, the current design has limitations in citation granularity. 
First, citations are limited to the sentence level, meaning that we do not distinguish between multiple claims within a single sentence. 
For example, if a sentence contains multiple claims supported by different evidence, we treat it as a full sentence-level citation.
Second, \method treats subfigures or subtables (e.g., Figure 1a, 1b) as part of the entire figure or table, without distinguishing between them.
These limitations highlight areas for future improvement in handling fine-grained attribution tasks.


\bibliography{anthology}
\bibstyle{acl_natbib}

\clearpage
\appendix
\newpage

\section{Prompt Design}
\label{app:prompt}

\subsection{Data Processing Prompts}
\label{app:data_processing_prompt}

We list the prompts used for extracting Explanation QA and generating Locating QA in Table~\ref{prompt:extract_prompt},~\ref{prompt:generate_qa_prompt}.

\begin{table*}[!h]
\begin{tcolorbox}
\textbf{Task Overview}

Your task is to extract valid question-answer-evidence (Q-A-E) triples from rebuttal sections of research papers on OpenReview. The extracted triples must meet the following criteria:

\textbf{Question:} Neutral, logically self-contained, and directly related to the paper’s content. The question must not contain explicit citations (e.g., ``Section 4.3'' or ``Figure 2'').

\textbf{Answer:} The author’s response must include explicit citations to the paper’s main body content (e.g., ``Section 4.3, Line 39'' or ``Figure 2, Figure 3'').

\textbf{Evidence:} Citations in the answer must be precise and clearly formatted. Multiple references should be separated by commas.

\textbf{Definitions}

\textbf{Question:} A neutral, logically self-contained inquiry related to the paper’s content. The question must:
Avoid references to specific sections, lines, figures, or tables (e.g., ``Can Section 4.3 be clarified?'' is invalid).
Focus on exploring or clarifying the main body of the paper, excluding appendices.

\textbf{Answer:} The full response provided by the authors, which must:
Contain explicit citations to the paper’s content (e.g., ``Section 4.3, Line 39'').
Exclude vague or general references such as ``General response'' or ``Discussion section.''

\textbf{Evidence:} Explicit numerical references from the author’s response, such as: ``Section 4.3, Line 39'' ``Figure 2, Figure 3'' ``Table 5''

Evidence must be precise and, if there are multiple references, they should be separated by commas.

\textbf{JSON Output Format}
\begin{verbatim}
{
    "qas": [
        {
            "question": "Extracted question text.",
            "answer": "Author's response text.",
            "evidence": "Specific reference to the paper"
        }
    ]
}
\end{verbatim}
\caption{Prompt for extracting explanation questions.}
\label{prompt:extract_prompt}
\end{tcolorbox}
\end{table*}

\begin{table*}[!htbp]
\begin{tcolorbox}
\textbf{Task Overview}  

You will be provided with a portion of an academic paper, including text, images, tables, etc.  
Based on this content, generate multiple multiple-choice questions, each with four answer options.  

\textbf{Requirements for Generating Questions:}  

\textbf{Grounding Questions from Text:}  

The question must be directly answerable based on the provided paragraph.  
Focus on extracting clear, specific, and factual details such as model performance, data, or numerical values mentioned in the text.  

\textit{Examples:}

- ``What is the accuracy of Llama3 on the MMLU dataset?''

- ``What is the main evaluation metric used for the models?''

- ``Which model showed the highest accuracy on the given test?''

- ``What value was reported as the accuracy of Llama3 in the study?''

\textbf{Simple, Fact-based Questions:}  

Questions should not require external reasoning or inference.  
They should be straightforward and based solely on the provided content, such as factual details (e.g., accuracy, performance, test results).  

\textit{Examples:}

- ``What is the accuracy of the Llama3 model on the MMLU benchmark?''

- ``What dataset was used to evaluate the performance of the models?''

- ``Which model had the lowest error rate?''

\textbf{Avoid Reference to External Context:}  
Do not refer to figures, tables, or external sections of the paper. The questions should rely solely on the provided paragraph or text. Ensure that all the information needed to answer the question is contained within the paragraph itself.  

\textit{Examples:}

- ``What is the performance of Llama3 on the MMLU dataset?'' (without referring to ``Table 1'' or ``Figure 3'')

- ``What is the reported training time for the model?''

\textbf{Ensure Directness and Clarity:}  

The question must be simple and directly related to the paragraph's content, ensuring the answer can be explicitly found in the text.  

\textit{Examples:}

``What performance metric is used to evaluate Model A?''

``What was the result for Model X on the validation set?''

``What is the reported accuracy for Model B?''

\textbf{Refusal Field Usage:}  

If the provided content does not contain enough information to generate a valid question, set the \textbf{Refusal} field to \textbf{True}.

If the question meets the requirements and can be answered directly from the given paragraph, set the \textbf{Refusal} field to \textbf{False}.

\textit{Examples:}

\textbf{Refusal: True} (If the paragraph does not contain any measurable data or clear information)

\textbf{Refusal: False} (If the question can be answered based on the paragraph’s content)

\caption{Prompt for generating locating questions.}
\label{prompt:generate_qa_prompt}
\end{tcolorbox}
\end{table*}

\subsection{Evaluation Metric Prompts}
We list the prompts used for evaluating citation recall, citation precision, and the accuracy of explanation and locating questions in Table~\ref{prompt:eval_citation_recall},~\ref{prompt:eval_citation_precision},~\ref{prompt:eval_explanation_questions_acc},~\ref{prompt:eval_locating_questions_acc}.

\label{app:evaluation_metric_prompt}

\begin{table*}[!htbp]
\begin{tcolorbox}
You are an expert in evaluating text quality. You will receive a statement from an AI assistant’s response based on a paper, along with a part from the document (which could be a text paragraph, image, or table). Your task is to carefully assess whether this statement is supported by the provided part. Please use the following scale to generate your rating:

0: No support — The statement is largely unrelated to the provided part (text, image, or table), or most key points in the statement do not align with the content of the part.

1: Partially supported — More than half of the content in the statement is supported by the part, but a small portion is either not mentioned or contradicts the part.

2: Fully supported — Most information in the statement is supported by or extracted from the part. This applies only to cases where the statement and the part are almost identical.

Ensure that you do not use any information or knowledge outside of the provided part when evaluating. Please return only the rating in JSON format, with 0, 1, or 2.

Statement: \textbf{\{sentence\}}
\caption{Prompt for evaluating citation recall.}
\label{prompt:eval_citation_recall}
\end{tcolorbox}
\end{table*}

\begin{table*}[!htbp]
\begin{tcolorbox}
You are an expert in evaluating text quality. You will receive a statement from an AI assistant’s response based on a paper, along with a part from the document (which could be a text paragraph, image, or table). Your task is to carefully assess whether the provided part contains some key information of the statement. Please use the following scale to generate your rating:

0: Unrelevant — The statement is almost unrelated to the provided part, or all key points of the statement are inconsistent with the the provided part.

1: Relevant — Some key points of the statement are supported by or extracted from the the provided part.

Ensure that you do not use any information or knowledge outside of the provided part when evaluating. Please return only the rating in JSON format, with 0 or 1.

Statement: \textbf{\{sentence\}}

\caption{Prompt for evaluating citation precision.}
\label{prompt:eval_citation_precision}
\end{tcolorbox}
\end{table*}

\begin{table*}[!htbp]
\begin{tcolorbox}
You are an assistant skilled in evaluating text quality.  
Please evaluate the quality of an AI assistant’s response to a reviewer’s question. Since the response is addressing a reviewer’s inquiry regarding a paper, you need to evaluate the answer from the following dimensions:

1. \textbf{Similarity with the Author's Response}

   - \textbf{Definition}: Evaluate how similar the model’s response is to the author's response in terms of content, specifically whether the model’s answer aligns with the key points and reasoning of the author’s reply.  

   - \textbf{Evaluation Criteria}: If the model’s response covers the main points of the author’s reply and is highly similar in content, score it higher; if the model’s response significantly differs from the author’s content, score it lower.

2. \textbf{Completeness of the Response}
   
   - \textbf{Definition}: Evaluate whether the model’s response covers all the points raised by the reviewer and fully addresses their question.  
   
   - \textbf{Evaluation Criteria}: If the model’s answer includes all key aspects raised by the reviewer and addresses the question comprehensively, score it higher; if the model misses important points or fails to address key aspects, score it lower.

3. \textbf{Logical Coherence}
   
   - \textbf{Definition}: Evaluate whether the model’s response has a clear logical structure and coherent reasoning.  
   
   - \textbf{Evaluation Criteria}: If the model’s response is logically sound and the reasoning is coherent, score it higher; if there are logical flaws or incoherent reasoning, score it lower.

4. \textbf{Clarity and Expression}
   
   - \textbf{Definition}: Evaluate whether the model’s response is concise, clear, and easy to understand, and if it matches the author's language style.  
   
   - \textbf{Evaluation Criteria}: If the model’s response is straightforward, logically clear, and aligns with the author’s style, score it higher; if the response is lengthy, hard to understand, or deviates from the author's language style, score it lower.

\textbf{Process:}

1. Compare the AI assistant's answer with the reference answer, and evaluate the AI’s response based on the above dimensions. After evaluating each dimension, provide a score.

2. Your scoring should be strict, and follow these guidelines:  

   - If the model’s response is irrelevant or generates harmful content, the total score must be 0.  
   
   - If the model’s response shows significant gaps compared to the reference answer or performs poorly in multiple dimensions, the score should be 1.  
   
   - If the model’s response is similar to the reference answer and performs well in all dimensions, the score should be 2.  
   
   - Please return your scores in JSON format.
\caption{Prompt for evaluating explanation questions}
\label{prompt:eval_explanation_questions_acc}
\end{tcolorbox}
\end{table*}

\begin{table*}[!htbp]
\begin{tcolorbox}
You are asked to evaluate the quality of the AI assistant’s answers to user question as an impartial judge, and your evaluation should take into account factors including correctness (high priority), and comprehensiveness (whether the assistant’s answer covers all points). Read the AI assistant’s answer and compare against the reference answer, and give an overall integer rating in 0, 1, 2 (0 = wrong or irrelevant, 1 = partially correct, 2 = absolutely correct) based on the above principles, strictly in the following format: {"answer\_rating": 2} (where 2 is just an example). So your JSON output must have the shape {"answer\_rating": <integer>}.
\caption{Prompt for evaluating locating questions}
\label{prompt:eval_locating_questions_acc}
\end{tcolorbox}
\end{table*}

\subsection{Source Identification Prompt}
\label{app:source_identification_prompt}
We list the prompt used to evaluate whether a model can identify the most relevant source for answering a given question in Table~\ref{prompt:source_attribution_mcq}.

\begin{table*}[!htbp]
\begin{tcolorbox}
Please identify the most relevant source of evidence to locate information that could address the following query. Provide your answer by selecting one of the options: A, B, C, D, or E. Begin your response with the selected letter and, if necessary, briefly explain why it is the most relevant source.

Identify where information about

\textbf{\{question\}}

can be found.
                        
Which source is most relevant?

\textbf{\{options\}}

\caption{Prompt used to evaluate source identification ability in the multi-choice setting.}
\label{prompt:source_attribution_mcq}
\end{tcolorbox}
\end{table*}

\section{Human Evaluation}
\subsection{Evidence Paring}
\begin{figure*}[h]
    \centering
    \includegraphics[width=\linewidth]{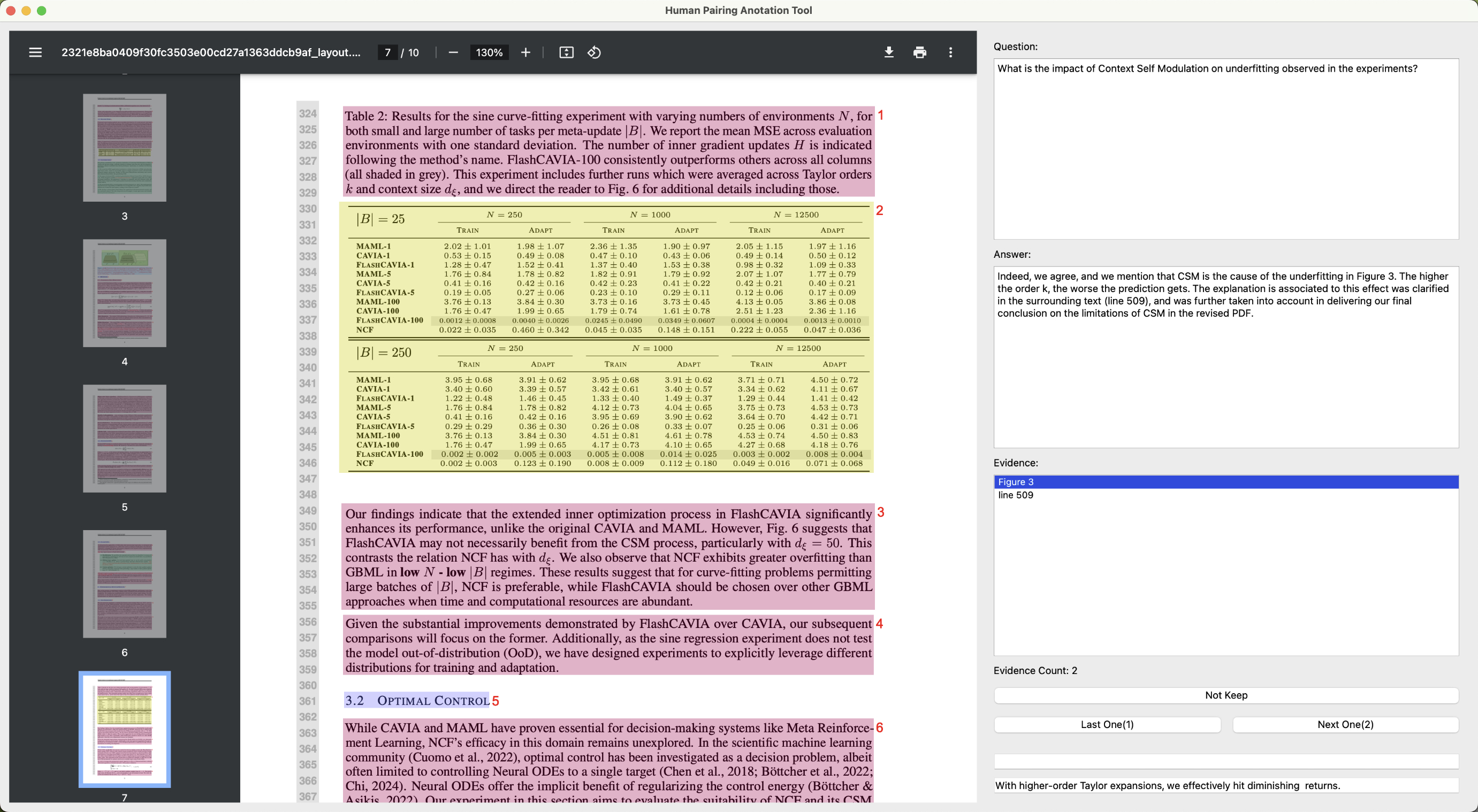}
    \caption{GUI screenshot for human annotators to map each reference to its corresponding content.}
    \label{fig:pairing_GUI}
\end{figure*}

Human annotators map each reference to its corresponding content using the GUI shown in Figure~\ref{fig:pairing_GUI}.

\subsection{Quality Control}
Our annotation process involves three students from the artificial intelligence field, with one serving as the annotation lead. The process takes approximately one month to complete, and annotators are compensated at the local minimum hourly wage rate. Regarding inter-annotator agreement, in cases of disagreement about whether to retain specific data points, the annotation lead makes the final decision.

\begin{figure*}[h]
    \centering
    \includegraphics[width=\linewidth]{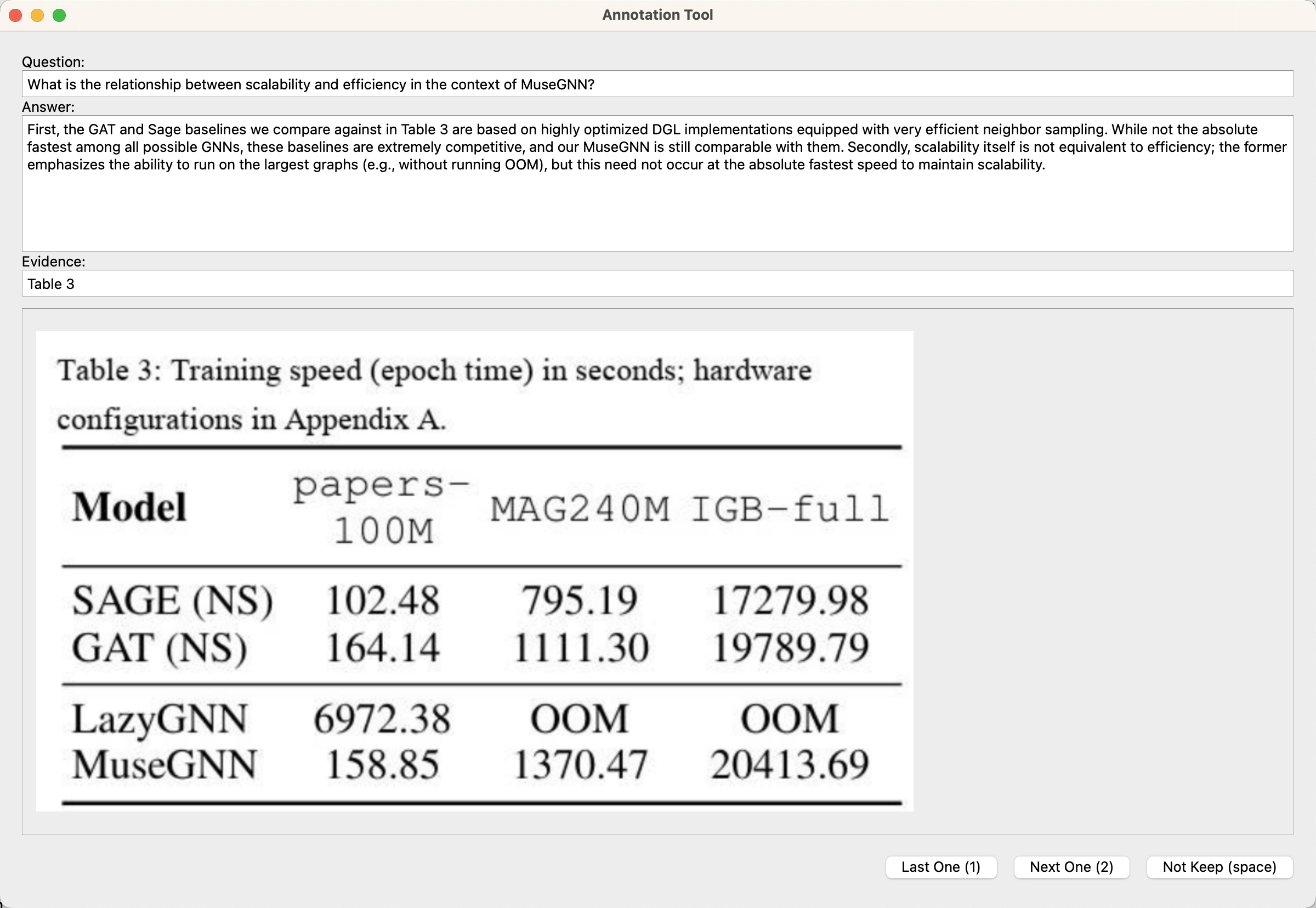}
    \caption{GUI screenshot for verifying filtered QA.}
    \label{fig:filter_GUI}
\end{figure*}

Human annotators verify data quality and filter out bad cases using the GUI shown in Figure~\ref{fig:filter_GUI}.

\label{app:quality_control}

\subsection{Agreement Between Human Annotations and GPT-4o}
\label{app:human_check}
\begin{figure*}[h]
    \centering
    \includegraphics[width=\linewidth]{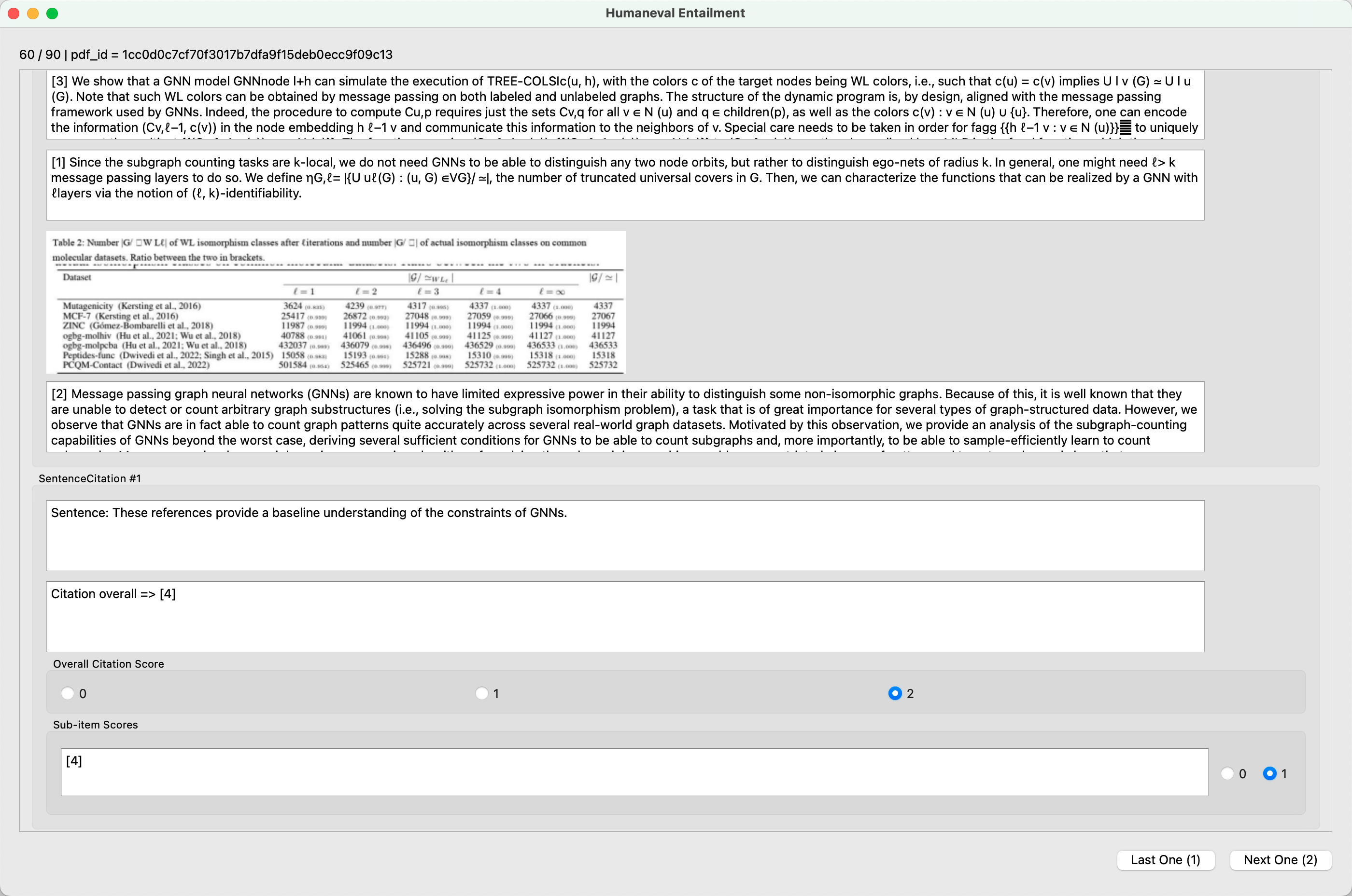}
    \caption{GUI screenshot for human-annotated entailment verification.}
    \label{fig:entailment_GUI}
\end{figure*}

To verify the accuracy of our evaluation pipeline, we conducted a manual annotation study on 75 model-generated responses, comprising 25 objective questions and 50 subjective questions, resulting in over 457 entailment judgments. 
We then compared these human annotations with the entailment judgments produced by GPT-4o. 
As shown in Table \ref{tab:human_check}, the results indicate a high degree of agreement between human annotations and GPT-4o's predictions, demonstrating the reliability and correctness of our pipeline.
The annotation GUI is shown in Figure~\ref{fig:entailment_GUI}.

\begin{table*}[t]
  \centering
  \small
  \begin{tabular}{lcccccc}
    \toprule
    \multirow{2}{*}{\textbf{Model}} 
    & \multicolumn{3}{c}{\textbf{Subjective}} 
    & \multicolumn{3}{c}{\textbf{Objective}} \\
    \cmidrule(lr){2-4} \cmidrule(lr){5-7}
    & \textbf{F1} & \textbf{Recall} & \textbf{Precision} 
    & \textbf{F1} & \textbf{Recall} & \textbf{Precision} \\
    \midrule
    GPT-4o      & 0.80 & 0.80 & 0.79 & 0.82 & 0.81 & 0.83 \\
    \bottomrule
  \end{tabular}
  \caption{Entailment Judgment Alignment: Model vs. Human Ground Truth}
  \label{tab:human_check}
\end{table*}

\section{Ablation Study}
\subsection{Effect of LLM Judge Choice}
\label{app:llm_judge}
To further investigate the robustness of our automatic evaluation setup, we conduct an ablation study using alternative judge models. Specifically, we evaluate model performance on 90 randomly sampled examples (30 \textit{Locating} and 60 \textit{Explanation} questions), comparing scores assigned by GPT-4o and DeepSeek V3-0324.

As shown in Table~\ref{tab:llm_judge_ablation}, GPT-4o consistently achieves the highest scores under both judge models. While DeepSeek V3-0324 tends to yield slightly higher absolute scores across all models, the relative ranking remains consistent. This suggests that self-preference bias from GPT-4o does not significantly affect evaluation outcomes, confirming the robustness of our LLM-based evaluation setup.

\begin{table}[t]
  \centering
  \small
  \begin{tabular}{lcc}
    \toprule
    \textbf{Model}
    & \textbf{GPT-4o Judge}
    & \textbf{DeepSeek V3 Judge} \\
    \midrule
    \textcolor{gray!50}{\textbf{Proprietary}} \\
    GPT-4o-2024-11-20 & 66.72 & 76.11 \\
    GPT-4o-mini       & 64.55 & 72.22 \\
    \textcolor{gray!50}{\textbf{Open-Source(7-14B)}} \\
    Qwen2-VL-7B-Instruct & 63.77 & 65.00 \\
    InternVL2\_5-8B      & 62.53 & 67.78 \\
    \bottomrule
  \end{tabular}
  \caption{Accuracy of citation evaluation across models under different LLM judges on a 90-sample subset.}
  \label{tab:llm_judge_ablation}
\end{table}

\subsection{Effect of Captions on Citation Performance}
\label{app:caption_ablation}

To assess the role of visual-textual information in multimodal citation understanding, we conduct a comprehensive ablation study across the full benchmark dataset (3,000 examples), comparing model performance with and without image captions.

As shown in Table~\ref{tab:caption_ablation}, we observe that including captions leads to minor changes in performance across most evaluation metrics. Notably, the accuracy improvements are modest for both GPT-4o-mini and GPT-4o. Interestingly, in some cases (e.g., GPT-4o-mini), the inclusion of captions slightly degrades performance in label prediction and citation generation (as measured by F1 and exact match), while GPT-4o exhibits a substantial gain in citation F1.

These results demonstrate that our benchmark does not rely solely on OCR-extracted text, and that the image-caption setting we adopt provides a reasonable and realistic testbed for evaluating MLLMs' citation capabilities. At the same time, the relatively limited gains from caption inclusion highlight that current models still face challenges in grounding their responses effectively, even when textual cues are explicitly embedded in the image.

\begin{table}[t]
  \centering
  \small
  \begin{tabular}{lcc}
    \toprule
    \textbf{Model} & \textbf{GPT-4o-mini} & \textbf{GPT-4o} \\
    \midrule
    No Cap. Acc & 62.25 & 64.92 \\
    With Cap. Acc & 64.55 & 66.72 \\
    \textbf{Acc Impr.} & \textbf{+2.30} & \textbf{+1.80} \\
    \midrule
    No Cap. S-F1 & 48.63 & 67.10 \\
    With Cap. S-F1 & 45.34 & 68.19 \\
    \textbf{S-F1 Impr.} & \textbf{-3.29} & \textbf{+1.09} \\
    \midrule
    No Cap. S-EM & 28.98 & 38.47 \\
    With Cap. S-EM & 25.34 & 39.11 \\
    \textbf{S-EM Impr.} & \textbf{-3.64} & \textbf{+0.64} \\
    \midrule
    No Cap. C-F1 & 55.51 & 76.54 \\
    With Cap. C-F1 & 49.53 & 87.43 \\
    \textbf{C-F1 Impr.} & \textbf{-5.98} & \textbf{+10.89} \\
    \bottomrule
  \end{tabular}
  \caption{Effect of captions on citation evaluation performance across multiple metrics. The performance with captions is compared to that without captions.}
  \label{tab:caption_ablation}
\end{table}

\end{document}